\documentclass[letterpaper, 10 pt, conference]{ieeeconf}  

\IEEEoverridecommandlockouts                              

\usepackage[l2tabu,orthodox]{nag}

\usepackage[
backend=bibtex8, 
style=ieee, 
sorting=none, 
natbib=true, 
doi=false, 
isbn=false, 
url=false, 
eprint=false, 
maxcitenames=1, 
mincitenames=1
]{biblatex}

\usepackage[draft, pdftex,colorlinks]{hyperref}

\usepackage{layouts} 

\usepackage[printonlyused]{acronym}

\usepackage{siunitx}
\usepackage[all]{nowidow}

\usepackage[dvipsnames, table]{xcolor}

\usepackage{lipsum}

\usepackage{verbatim}

\usepackage{pdflscape}

\usepackage{pifont}
\newcommand{\cmark}{\ding{51}}
\newcommand{\xmark}{\ding{55}}

\usepackage[pdftex]{graphicx}
\usepackage{epstopdf}

\usepackage{import}

\usepackage{float}

\usepackage[font=footnotesize]{caption}

\graphicspath{{./latexGoodPractices/}}

\usepackage{booktabs}

\usepackage{tabu}

\usepackage{supertabular}

\usepackage{rotating,tabularx}

\usepackage{makecell}

\usepackage{amssymb,amsfonts,amsmath,amscd}

\usepackage{bm} 

\newcommand{\norm}[1]{\left\Vert#1\right\Vert}

\newcommand{\bbm}{\begin{bmatrix}}
\newcommand{\ebm}{\end{bmatrix}}

\usepackage{enumerate}
\usepackage[normalem]{ulem}

\acrodef{ICP}{Iterative Closest Point} 
\acrodef{RANSAC}{Random Sample Consensus}
\acrodef{LTS}{least-trimmed squares}
\acrodef{LIDAR}{Light Detection And Ranging}
\acrodef{FOV}{Field Of View}
\acrodef{FGR}{Fast Global Registration}
\acrodef{FPFH}{Fast Point Feature Histogram}
\acrodef{FRMSD}{Fractional Root Mean Squared Distance}
\acrodef{RQE}{R\'{e}nyi Quadratic Entropy}
\acrodef{KC}{Kernel Correlation}
\acrodef{TrICP}{trimmed distance outlier filter}
\acrodef{L2}{squared distance}
\acrodef{AICP}{Auto-tuned ICP}
\acrodef{LM-ICP}{Levenberg-Marquardt ICP}
\acrodef{EM-ICP}{Expectation Maximization ICP}
\acrodef{RMT}{Relative Motion Threshold}
\acrodef{NDT}{Normal Distribution Transformation}
\acrodef{GM}{Geman-McClure}
\acrodef{GICP}{Generalized-ICP} 
\acrodef{IRLS}{Iteratively Reweighted Least-Squares}
\acrodef{MAD}{Median of Absolute Deviation}
\acrodef{SLAM}{Simultaneous Localization and Mapping}

\title{\LARGE \bf
Analysis of Robust Functions for Registration Algorithms
}

\author{Philippe Babin, Philippe Gigu\`{e}re and Fran\c{c}ois Pomerleau$^{*}$%
\thanks{$^{*}$The authors are from the Northern Robotics Laboratory, Universit\'{e} Laval, Canada.	\newline
        {\tt\small \{philippe.babin.1@, philippe.giguere@ift.,\newline
        	francois.pomerleau@ift.\}ulaval.ca }}
}

\begin{document}

\maketitle
\thispagestyle{empty}
\pagestyle{empty}


\begin{abstract}

Registration accuracy is influenced by the presence of outliers and numerous robust solutions have been developed over the years to mitigate their effect.
However, without a large scale comparison of solutions to filter outliers, it is becoming tedious to select an appropriate algorithm for a given application. 
This paper presents a comprehensive analyses of the effects of outlier filters on the \ac{ICP} algorithm aimed at  mobile robotic application.
Fourteen of the most common outlier filters (such as M-estimators) have been tested in different types of environments, for a total of more than two million registrations.
Furthermore, the influence of tuning parameters have been  thoroughly explored.
The experimental results show that most outlier filters have similar performance if they are correctly tuned.
Nonetheless, filters such as  \emph{Var. Trim.}, \emph{Cauchy}, and \emph{Cauchy MAD} are more stable against different environment types.
Interestingly, the simple norm $L_1$ produces comparable accuracy, while been parameterless.

\end{abstract}

%
%
\vspace*{-5pt}
\section{Introduction}
A fundamental task in robotics is finding the rigid transformation between two overlapping point clouds.
The most common solution to the point cloud registration problem is the \ac{ICP} algorithm, which alternates between finding the best correspondence for the two point clouds and minimizing the distance between those correspondences~\citep{Besl1988,Chen1992}.
Based on the taxonomy of \citet{Rusinkiewicz2001}, \citet{Pomerleau2013a} proposed a protocol and a framework to test and compare the common configurations of \ac{ICP}.
They simplified the process to four stages: 1) data point filtering, 2) data association, 3) outlier filtering, and 4) error minimization.
The stage on outlier filtering is necessary as the presence of a single outlier with a large enough error could have more influence on the minimization outcome than all the inliers combined.
To solve this generic problem, \citet{Huber1964} extended classical statistics with robust cost functions.
A robust cost function reduces the influence of outliers in the minimization process.
The most common class of these robust functions is the maximum likelihood estimator, or M-estimator. Other solutions exist, which rely either on thresholds (i.e., hard rejection) or on continuous functions (i.e., soft rejection).

\begin{figure}
\centering
\includegraphics[width=\columnwidth]{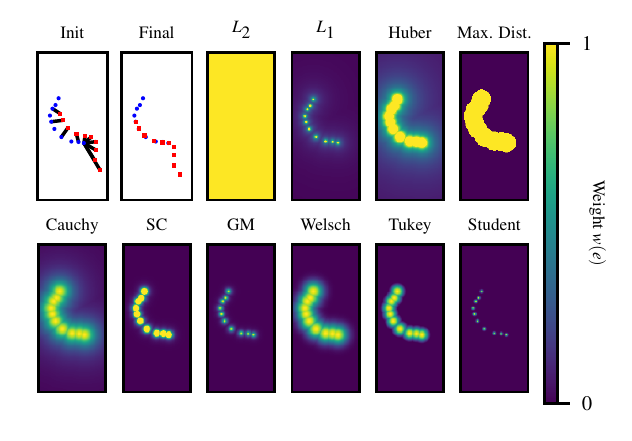}
\caption{
	The effect of different outlier filters on the registration of two overlapping curves. 
	The blue circles and red squares are respectively the reference points and the reading points.
	The initial graph depicts the data association from the nearest neighbor at the first iteration.
	The final graph shows the registration at the last iteration.
	The other graphs depict  scalar fields of the weight function $w(e)$ of a reading point.
	For $L_2$, all points have the same weight.
	$L_1$ gives an infinite weight to a point directly on top of a reference point.
}
\vspace*{-20pt}
\label{fig:overview}
\end{figure}

To the best of our knowledge, the current research literature is missing a comprehensive comparison of outlier filters for \ac{ICP} (i.e., Stage 3 of the \ac{ICP} pipeline).
In the case of \ac{ICP} used in mobile robotics, outliers are mainly caused by non-overlapping regions, sensor noises and shadow points produced by the sensor.
Most papers about outlier filters compare their own algorithm with only two other algorithms~\citep{Fallon2017,Granger2002,Fitzgibbon2002} or only on a single dataset~\citep{Pomerleau2010,Bosse2016,Granger2002,Fitzgibbon2002}.
Few papers evaluate the influence of the overlap between the two point clouds and the initial perturbation, which are leading error causes for \ac{ICP}~\cite{Pomerleau2015}. 
Furthermore, the dataset selected for evaluation varies depending on research fields. 
For instance, papers targeting object reconstruction will use the Stanford dataset~\citep{Phillips2007, Fitzgibbon2002, Bouaziz2013a}. 
Results obtained with dataset containing exclusively objects might be too specific and consequently not translate well to the field of mobile robotics, because of the difference in structure, density, and scale.
Also, experiments on registration performances tend to modify multiple stages of \ac{ICP} at once, making it difficult to estimate the impact of outlier rejection algorithms on the overall performance.
Additionally, few papers on outlier filter evaluate the impact of the tuning parameters within those outlier rejection algorithms. 
The influence of an outlier for a given error can change drastically, as a function of the tuning parameter value.
Finally, with the rise of the number of \ac{ICP} variants~\cite{Pomerleau2015}, it is becoming tedious to select the appropriate robust solution for an application in mobile robotics.

%
%
To mitigate those problems, we propose a comprehensive analysis of outlier filter algorithms.
To this effect, the main contributions of this paper are:
1) a large scale analysis investigating 14 outlier filters subject to more than two millions registrations in different types of environment;
2) a consolidation of the notion of point cloud \emph{crispness}~\cite{Sheehan2013} to a special case of a common M-estimator, leading to a better understanding of its use in the outlier filtering stage; 
3) a support to better replication of our results with the open-source implementations of tested outlier filters in  \texttt{libpointmatcher}\footnote{\url{https://github.com/ethz-asl/libpointmatcher}}.

\vspace*{-5pt}
\section{Related Works}
\vspace*{-5pt}
\label{related_works}
%
%
%

Prior to the existence of \ac{ICP}, a seminal work on M-estimators was realized by \citet{Welsch1977}, where he surveyed eight functions now known as the classic M-estimators.
In the context of camera-based localization, \citet{MacTavish2015} did a thorough effort by comparing seven robust cost functions, but their conclusions do not translate directly to point cloud registrations.
Closer to us, \citet{Pomerleau2015} proposed an in-depth review of the literature explaining how outlier filters must be configured depending on the robotic application at hand.
Unfortunately, their investigation is limited to listing current solutions, without comparison.
We took inspiration from those surveys to select our list of solutions analyzed in this paper.

When it comes to comparing outlier filters, the most common baseline is vanilla \ac{ICP} (i.e., labeled \emph{$L_2$} hereafter), which does not have any outlier filter and directly minimize a least-squared function.

In terms of hard rejections, \citet{Phillips2007} compared a solution using an adaptive trimmed solution against  a manually adjusted trimmed threshold~\cite{Rusinkiewicz2001} and $L_2$.
However, they limited their analysis to pair of point clouds with overlap larger than 75\% and using simulated outliers.
%
Another adaptive threshold solution, this time related to simulated annealing algorithms~\cite{Pomerleau2010}, was compared to five types of hard rejection algorithms.
Unfortunately, they used a limited 2D dataset relying on ten pairs of scans with similar overlap.
As for soft rejections, \citet{Bergstrom2014} compared the effect of three M-estimators  on \ac{ICP} and proposed an algorithm to auto-tune them, but they only provided results based on simulated data of simple geometric shapes.
%
\citet{Agamennoni2016} proposed a soft rejection function based of Student's T-distribution for registration between a sparse and a dense point clouds.
But, they compare their algorithm to another complete \ac{ICP} solution, where multiple stages changed.
%
Closer to the topic of our analysis, \citet{Bosse2016} compared five M-estimators and another robust function against $L_2$, for a variety of minimization problem, one of which was \ac{ICP}.
However, the \ac{ICP} analysis was limited to examples relying on a single 2D pair of point clouds.
In this analysis, a common 3D benchmark is used for all solutions allowing a fair comparison of robust cost functions.

%
%

When an outlier filter rely on fixed parameters, it is important to know their effects on registration performance.
\citet{Segal2009} removed matching points with residual errors higher than a given value (i.e., named \emph{max. distance} hereafter) and evaluated the influence of this parameter on \ac{ICP}.
They concluded that the value used for this parameter is a trade-off between accuracy and robustness. 
\citet{Bosse2008} compared the same outlier filter against one of the classic M-estimator \emph{Cauchy} in an 2D outdoor environment.
They concluded that \emph{max. distance} is less robust to the parameter value than \emph{Cauchy}, but it provided better accuracy when correctly tuned.

For the most part, a robust cost function relies on fix parameters, configured by trial and error.
To sidestep this issue, some outlier filters are designed so as to be auto-tunable. 
For instance, \citet{Haralick1989} and \citet{Bergstrom2014} have both proposed an algorithm to auto-scale M-estimators.
However, \citet{Bergstrom2014} requires two additional hyper-parameters to tune the auto-scaler: one representing the sensor's standard deviation and the other specifying the decreasing rate to reach the standard deviation. 
In the case of outlier filters based on a threshold following a given quantile of the residual error, \citet{Chetverikov2005} proposed an estimator to tune the overlap parameter, which use an iterative minimizer based on the Golden Section Search algorithm.
Unfortunately, having an iterative minimizer inside \ac{ICP}'s iterative loop becomes computationally intensive~\citep{Du2011}, thus most of the implementation resort to manually fixing the quartile.
\citet{Phillips2007} improved on \citep{Chetverikov2005} by minimizing the \ac{FRMSD} to tune the parameter representing the overlap ratio.
It removes the need for an inner loop to estimate the parameter, thus speeding up the execution by a factor of at least five.
But, this algorithm performance also depends on a hyper-parameter that needs to be tuned~\citep{Du2011}.
In this paper, we also explore the influence of tuning parameters, testing even the stability of hyper-parameters, and provide a methodology to tune them.


\vspace*{-5pt}
\section{Theory}
\vspace*{-5pt}
\label{theory}
%
%

The \ac{ICP} algorithm aims at estimating a rigid transformation $\hat{\bm{T}}$ that best align a \emph{reference} point cloud $\mathcal{Q}$ with a \emph{reading} point cloud $\mathcal{P}$, given a prior transformation $\check{\bm{T}}$. 
The outlier filtering stage of \ac{ICP} has strong ties to error minimization.
The former reduces the influence of wrongful data association, while the latter finds a solution that respects the constraints of the previous stage. In the context of \ac{ICP}, these two stages can be summarized as estimating the rigid transformation $\hat{\bm{T}}$ by minimizing
\begin{equation}
\label{eq:icp_min}
\hat{\bm{T}} = \arg \min_{\bm{T}}  \sum_{i=1} \sum_{j=1} \rho \Big( e(\bm{T}, \bm{p_i}, \bm{q_j}) \Big),
\vspace*{-5pt}
\end{equation}
where $\rho(\cdot)$ is the cost function.
The error function $e(\cdot)$ is the scaled distance between matched points, defined as
\begin{equation}
\label{eq:error}
e(\bm{T}, \bm{p_i}, \bm{q_j}) = \frac{\norm{\bm{T}\bm{p_i} - \bm{q_j}}}{s},
\vspace*{-5pt}
\end{equation}
where $\bm{p_i} \in \mathcal{P}$ and $\bm{q_j} \in \mathcal{Q}$.
This is equivalent to a simplified version of the Mahalanobis distance, where the scale $s$ is uniform on all dimensions.
For the original version of \ac{ICP}, $s$ is set to one.
The double summation of \autoref{eq:icp_min} is expensive to compute and is typically approximated using a subset of pairs, using nearest neighbor points of each $\bm{p_i}$.
To simplify the notation, we will use $e_m(\cdot)$ for each error to be minimized, with $m$ being the index of this subset.
Since $\rho(\cdot)$ is non-convex, the minimization must be solved iteratively, in a manner similar to the \ac{IRLS}~\citep{Bergstrom2014} by decomposing $\rho(\cdot)$ as a weight $w$ and a squared error term, such that
\begin{equation}
\label{eq:icp_min} 
\hat{\bm{T}} \approx \arg\min_{\bm{T}}  \sum_{m=1} w \big ( e_m(\check{\bm{T}}) \big ) \ e_m(\bm{T})^2 .
\vspace*{-5pt}
\end{equation}
At each iteration, the prior transformation $\hat{\bm{T}}$ is assigned to the last estimated transformation until convergence. 
\autoref{fig:overview} shows a toy example of registration. 
The subsequent scalar fields show the impact of different weight functions $w(\cdot)$ in the neighborhood of $\mathcal{Q}$.
A notable example is $L_2$, where the lack of weights is equivalent to using a constant weight for all pairs.
Beyond $L_2$, outlier filtering is all about the choice of this weight function $w(\cdot)$ and its configuration.
\autoref{tab:m_estimators} shows a list of outlier functions used in this paper, with a dedicated column highlighting their implementation of $w(\cdot)$.


\begin{table}[htbp]
   \vspace{-5pt}
	\setlength\tabcolsep{1pt}
	\centering
    	\caption{%
		Descriptive table of robust cost functions used in this analysis expressed with respect of their tuning parameter $k$ and the scaled error $e$.
	}
    \label{tab:m_estimators}
	\taburowcolors[2]1{white..gray!15}
\begin{tabu} {Xllll}
\toprule
Functions &
Conditions & 
Cost  $\rho(e)$ &
Weight $w(e)$ &
M
\\
\midrule
$L_2$&
&
$\frac{e^2}{2}$&
$1$&
\cmark 
\\
$L_1$&
&
$|e|$&
$\frac{1}{|e|}$& 
\xmark 
\\ 
Huber&
$
\left\{
\begin{array}{l}
	|e| \leq k  \\
	\text{otherwise}
\end{array}
\right.
$
&
$
\left\{
\begin{array}{l}
\frac{e^2}{2}  \\
k(|e| -k/2)
\end{array}
\right.
$
&
$
\left\{
\begin{array}{l}
1  \\
\frac{k}{|e|}
\end{array}
\right.
$
&
\cmark 
\\ 
Cauchy&
&
$\frac{k^2}{2}\log(1+(e/k)^2)$&
$\frac{1}{1 + (e/k)^2}$&
\cmark 
\\ 
GM&
&
$\frac{e^2/2}{k+e^2}$&
$\frac{k^2}{(k + e^2)^2}$&
\cmark 
\\ 
SC&
$
\bigg\{\begin{array}{lr}
e^2 \leq k  \\
\text{otherwise}
\end{array}$
&
$
\bigg\{\begin{array}{lr}
\frac{e^2}{2}  \\
\frac{2 k e^2}{k+e^2} -k/2
\end{array}$
&
$\bigg\{\begin{array}{lr}
1\\
\frac{4 k^2}{(k+e^2)^2}
\end{array}$
&
\cmark 
\\ 
Welsch &
&
$\frac{k^2}{2}(1-\exp(-(\frac{e}{k})^2))$&
$\exp(-(e/k)^2)$&
\cmark 
\\ 
Tukey &
$
\bigg\{\begin{array}{lr}
|e| \leq k  \\
\text{otherwise}
\end{array}$
&
$
\bigg\{\begin{array}{lr}
\frac{k^2(1-(1-(\frac{e}{k})^2)^3)}{2} \\
\frac{k^2}{2}  
\end{array}$
&
$
\bigg\{\begin{array}{lr}
(1-(e/k)^2)^2 \\
0
\end{array}$
&
\cmark 
\\ 
Student &
&
&
$
\frac{(k + 3)(1 + \frac{e^2}{k})^{-\frac{k + 3}{2}}}{k + e^2}
$
&
\xmark 
\\ 
Max. Dist. &
$\bigg\{\begin{array}{lr}
|e| \leq k  \\
\text{otherwise}
\end{array}$
&
$\bigg\{\begin{array}{lr}
\frac{e^2}{2}  \\
\frac{k^2}{2}  
\end{array}$
&
$\bigg\{\begin{array}{lr}
1\\
0
\end{array}$
&
\cmark
\\ 
Trimmed
&
\multicolumn{2}{l}{
$\bigg\{\begin{array}{lr}
e \leq P_f 
\\
\text{otherwise}
\end{array}$
}
&

$\bigg\{\begin{array}{lr}
1 \\
0
\end{array}$
&
\xmark 
\\
\bottomrule

\end{tabu}
    \vspace{2pt}
	\footnotesize{Legend: M = Is the function a M-Estimator?}
   \vspace{-17pt}
\end{table}

\subsection{Hard rejection}
Outlier filters categorized as hard rejection define the result of $w(\cdot)$ to be binary (i.e., either zero or one).
The two most commons solutions are \emph{Max. distance} and \emph{Trimmed}.
The solution \emph{Max. distance} rejects any match with a distance larger than a threshold.
\emph{Trimmed} only keeps the error bellow the $f$\textsuperscript{th} percentile $P_f$ of the matches, where $f$ is the \emph{overlap ratio parameter}. This makes the registration accuracy directly related to how close this parameter $f$ is to the actual overlap between reference and reading. In that sense, if $f$ deviates from the true overlap, the accuracy will degrade.
In applications where the overlap is unknown or change often, selecting a fix overlap ratio $f$ becomes challenging.
A variant of \emph{trimmed}, \emph{Var. Trimmed} \citep{Phillips2007}, calculates the \ac{FRMSD} for all possible overlap ratios, and select the ratio with the minimum \ac{FRMSD} value as $f$.
The \emph{Median filter} is also used, 
but is a special case of \emph{Trimmed}, where $f = 50\,\%$.

\subsection{Soft Rejection and M-estimators}
Outlier filters using soft rejection output a weight where $w(\cdot) \in \mathbb{R}^+$.
The most common soft rejection algorithms are M-estimators.
To be an M-estimator, a cost function $\rho(\cdot)$ must fulfill three conditions, which are to be \emph{1)} symmetric, \emph{2)} non-negative, and \emph{3)} monotonically-increasing~\citep{Haralick1989}.
Those conditions do not limit M-estimators to robust cost function as, for example, $L_2$ satisfies all of them.
Moreover, cost functions associated to M-estimators are analyzed using their influence function $\psi(\cdot)$ and  weight function $w(\cdot)$, such that
\begin{align*}
\psi(e) = \frac{\partial \rho(e)}{\partial e}
& 
\qquad \text{and}
&
w(e) = \frac{\psi(e)}{e}.
\end{align*}
The influence function $\psi(\cdot)$ is used to evaluate whether an M-estimator is robust or not.
If $\psi(\cdot)$ is non-monotonic (i.e., redescending) and is null for an error that tend to infinity, the M-estimator is considered robust.
As for the weight function $w(\cdot)$, it is given for convenience since it is the only part required to implement an M-estimator for an \ac{IRLS} solution, as in \autoref{eq:icp_min}.
In the soft rejection algorithms considered in this paper (shown in \autoref{tab:m_estimators}), two are not M-estimators:
$L_1$ has a singularity for an error that is equal to zero, and \emph{Student} has an undefined $\rho(\cdot)$.

It is worth noting that some soft rejection functions are related. For instance,
\emph{Huber} uses a parameter $k$ to combine $L_1$ and $L_2$, in order to avoid the singularity at $e=0$ of $L_1$. 
Also, \emph{Switchable-Constraint} (labeled \emph{SC} hereafter), typically used in pose graph \ac{SLAM}, was expressed as a combination of $L_2$ and \emph{Geman-McClure} (labeled \emph{GM} hereafter) using a parameter $k$.
It shares the same cost function as Dynamic Covariance Scaling~\citep{MacTavish2015}.
\emph{SC} is expected to have similar results to \emph{GM} for extreme values of $k$.

\subsection{Estimating the scale}
As expressed in \autoref{eq:error}, we used a scaled error in our \ac{ICP} implementation.
Contrary to the filter parameter $k$, which should be globally constant, the scale $s$ is related to the point clouds and can be either fixed or estimated at every iteration. 
The scale $s$ relates to the uncertainty for which paired points with a certain error should be considered as outliers.
 There are multiple estimators for the scale, two of the most interesting are: 
1) \citet{Haralick1989} used the \ac{MAD} as a scale estimator and calculated it at each iteration;
2) \citet{Bergstrom2014} starts with $s=1.9 \cdot \mathrm{median}(e)$ and then gradually decrease $s$ at each iteration, to asymptotically reach a standard deviation $\sigma_*$. The parameter $\xi$ controls the convergence rate. 

\subsection{Relating \emph{crispness} to the M-estimator Welsch}
\label{ssec:crisp}
The notion of \emph{crispness} as a measure of how well two point clouds are aligned was introduced by \citet{Sheehan2012} in the context of sensor calibration.
It originate from the use of a Gaussian kernel in a measurement of \ac{RQE} for a kernel correlation approach to registration~\citep{Kanade2004}.
\ac{RQE} has the following cost function: 
\begin{equation}
\rho_\text{rqe}(e_*) =\exp{\left(-\frac{e_*^2}{4\sigma^2} \right)},
\label{eq:rqe_cost}
\end{equation}
where $e_*$ is the unscaled error and $\sigma$ is a tuning parameter.
\ac{RQE} has been described as an M-estimator by \citep{Kanade2004}, however, it has not related to an existing M-estimator.
This cost function is a special case of \emph{Welsch} with $k = 2$ and $s = \sigma$.
It means minimizing for \ac{RQE} is the same thing as using a \emph{Welsch} M-estimator with this configuration.
Thus, this configuration is expected to have good accuracy.



\vspace*{-5pt}
\section{Experiments}
\vspace*{-10pt}
\label{experiments}

Our study focuses on registration-based localization in the context of mobile robotics.
Our analysis follows a similar methodology as in \citet{Pomerleau2013a}, with a strict focus on changing the outlier filtering stage of \ac{ICP}. The other stages were kept the same.
\autoref{tab:icp_config} describes the \ac{ICP} pipeline used.

A number of key factors were selected to be tested jointly, in order to determine their influence. These factors were: \textbf{1)} the \textbf{outlier filter}, selected based on their popularity and their interesting properties (total of 12); \textbf{2)} the \textbf{configuration parameter} of the filters; \textbf{3)} the \textbf{environment} type (indoor, outdoor, etc.); and \textbf{4)} the \textbf{overlap} between the reference and reading scans.
Since \ac{ICP} is sensitive to the initial estimate, 128 registrations were computed with a random perturbation from the ground truth for each factor selection above. How some of these factors were sampled during experiments is detailed below.

\textbf{Configuration parameter sampling} Filter parameter values were sampled in a way to ensure efficient exploration of their configuration space. For instance, the tuning parameters $k$ of all M-estimators (i.e., \emph{Huber}, \emph{Cauchy}, \emph{SC}, \emph{GM}, \emph{Welsch} and \emph{Tukey}) and \emph{Student} were sampled evenly on a log scale. 
Two sampling value regions were defined. The first region was $Z_1 \in \big[\num{1e-6}, 0.1\big[$ and the second one was $Z_2 \in \big[0.1,~100\big]$.
$Z_1$ explores the asymptotic behavior of the algorithm for near-zero $k$ values, while $Z_2$ is where the parameter with the best accuracy is located for most estimators.
They are sampled 20 times for $Z_1$ and 30 times for $Z_2$.
For the error scale $s$ for M-estimator (see Eq.~\ref{eq:error}), two auto-scalers (\emph{Berg.} and \ac{MAD}) and one fix scale have been tested. 
All three permutations were tested on \emph{Cauchy}, while all other M-estimators were tested only with MAD\@.
In the case of the \emph{Berg.} auto-scaler, we used a convergence rate of $\xi = 0.85$, as it was found to be the best one in our analysis.
Its parameter $\sigma_*$ was sampled in the same two zones as the M-estimator ($Z_1$ and $Z_2$). \emph{Cauchy Berg.} used the tuning parameter $k_{cauchy} = 4.304$, and it was selected based on \citet{Bergstrom2014}. 
If the estimator used was \ac{MAD}, then $s = MAD(\bm{e})$,  otherwise $s = 1$.
In the case of the \emph{Trimmed} filter, its overlap parameter $f$ has been sampled linearly 20 times in the range $\big[\num{1e-4}, 100\big]\%$. 
For the \emph{Var. Trim.} filter, the minimum and maximum overlap parameters have been set to 40\% and 100\% respectively, since this is the minimum and maximum overlap of the dataset. 
Its $\lambda$ parameters has been sample linearly 20 times between 0.8 and 5. 
The filter \emph{Max. Distance} has been sampled 20 times linearly in the range $\big[0.1, 2\big]$. 
Finally, the $L_1$ and $L_2$ filters were used as-is, as they do not have any parameters.

\textbf{Environments}
Experiments were performed on the \emph{Challenging Datasets for Point Cloud Registration}~\citep{Pomerleau2012}. 
These provide a ground truth with \emph{mm}-level of precision.
Our analysis used 3 sets of point clouds from this dataset, one per type of environment: structured (\emph{hauptgebaude}), semi-structured (\emph{gazebo summer}) and unstructured (\emph{wood summer}).
As each of them contains around 35 point clouds, it allows a fine control of the ratio of overlap between point clouds. 
For each set, 12 pairs of point clouds where selected, to uniformly sample the overlap between 40\% and 100\%.

\begin{table}
	\centering
	\caption{
		Listing of the configuration of \texttt{libpointmatcher} used for these experiments.
	}
	\begin{tabu} {@{}llX@{}} 
		\toprule 
		Stage & Configuration & Description \\
		\midrule 
		Data association          &  \texttt{KDTree} & Three matches per point \\
		Data filtering  &  \texttt{SurfaceNormal} &  Density with 20 neighbors \\
		                 &  \texttt{MaxDensity} & Limit density to 10k pts/\si{\cubic\metre} \\
		 &  \texttt{RandomSampling} & Keep 75\% of points \\
		Error Min. & \texttt{PointToPlane} &  Point-to-plane error\\
		Trans. Checking & \texttt{Differential} & Stop below \SI{1}{mm} and \SI{1}{mrad}\\
	                        	& \texttt{Counter} & Max. iteration count is \num{40} \\
		\bottomrule 
	\end{tabu}
	\label{tab:icp_config}
	\vspace*{-17pt}
\end{table}

\textbf{Initial perturbation on $\bm{T}_0$}
For fine registration, \ac{ICP} requires a prior on the transformation between the \emph{reading} and \emph{reference} point clouds. 
The performance accuracy of \ac{ICP} is directly impacted by the distance between the initial transformation $\bm{T}_0$ and the ground truth. 
For our test, this initial transformation $\bm{T}_0$ was generated by adding a random perturbation sampled from a uniform distribution, and centered at the ground truth.
To stress-test \ac{ICP}, we choose a perturbation as challenging as the hard perturbation of \citep{Pomerleau2013a}.
The perturbation in translation was generated by sampling a point in a sphere with a \SI{1}{\meter} radius, while the one in rotation was generated by first sampling from an uniform angle distribution between 0 and \SI{25}{\deg} and then applying it around a random 3D vector. 

Our evaluation of accuracy used the transformation error $\bm{\Delta}$ defined as $\bm{\Delta} = \bm{T}_{gt}^{-1}\bm{T}_{final}$
where $\bm{T}_{gt}$ is the ground truth and $\bm{T}_{final}$ is the transformation at the last iteration.
The transformation error is further separated into two components, for easier interpretation:  $\bm{\Delta}_R$ for the 3x3 rotation matrix and $\bm{\Delta}_T$ for the 3x1 translation vector.
Finally, the translation error was evaluated with the euclidean distance of $\bm{\Delta}_T$, and 
the rotation error metric is $\theta=\arccos(\frac{trace(\bm{\Delta}_R) -1}{2})$.


\vspace*{-5pt}
\section{Results}
\vspace*{-1pt}
\label{results}
\begin{figure*}[htbp]
	\centering
	\includegraphics[width=\textwidth]{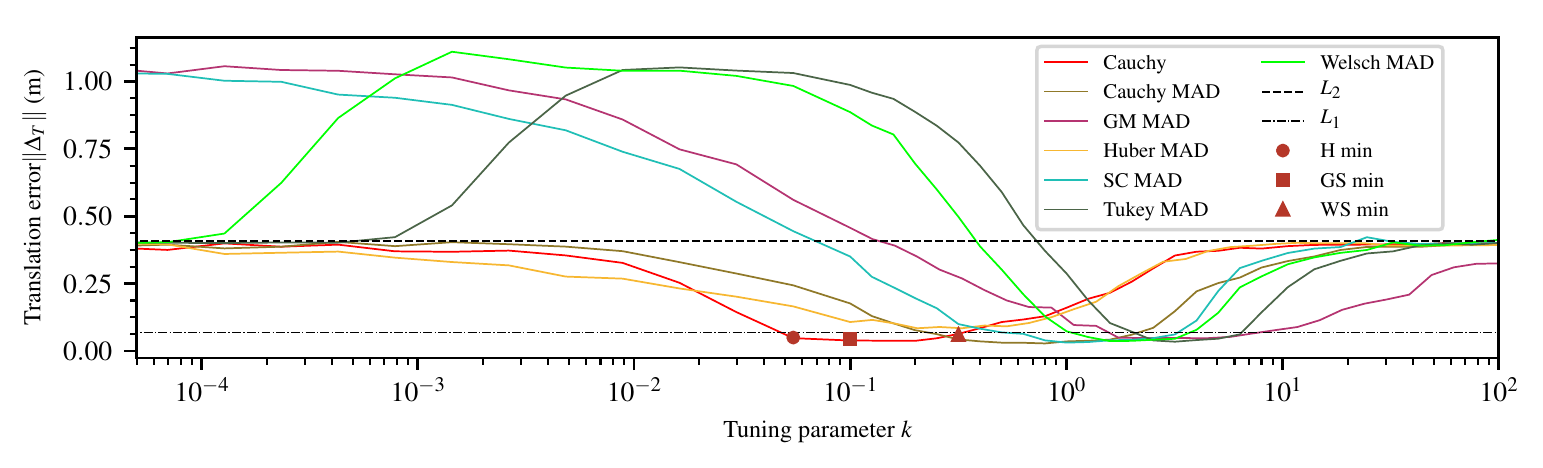}
	\vspace*{-20pt}
	\caption{
		Influence of the parameter's value on the translation registration accuracy of M-estimators.
        Each parameter sample point is the median of the error for all datasets for all overlap.
        The H min, GS min and WS min correspond to the parameter of \emph{Cauchy} with the minimum median error  for the environment \emph{Hauptgebaude}, \emph{Gazebo Summer}, and \emph{Wood Summer}.
}
	\label{fig:param_infl}
	\vspace*{-18pt}
\end{figure*}

\begin{figure}
	\centering
	\vspace*{-7pt}
	\includegraphics[trim={0 0 0 5cm},clip,width=0.47\textwidth]{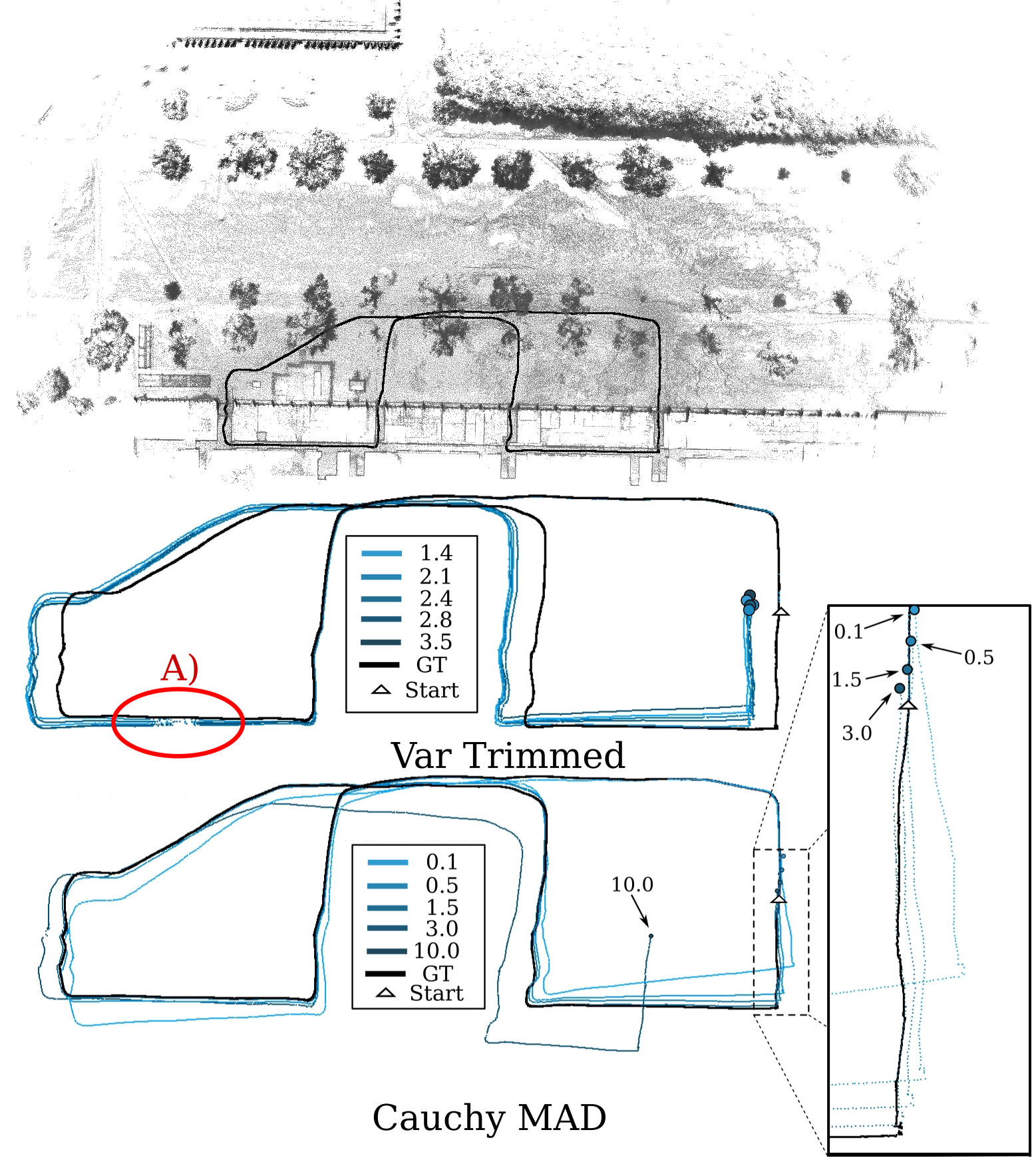}
	\vspace*{-7pt}
	\caption{%
		Testing the \ac{ICP} accuracy in a real time 3D SLAM application with a Husky A200 following a challenging indoor-outdoor route on Universit\'{e} Laval's campus.
		The start and end of the route are at the same localization.
		The black trajectory (ground truth) was calculated offline. 
		The end position of the trajectory for each parameter value is represented by a circle.
		In A), all configuration of \emph{Var. Trimmed} could not converge correctly in the corridor.}
	\label{fig:husky}
	\vspace*{-21pt}
\end{figure}

In the first set of tests (\autoref{sec:registration}), we performed over 2.3~million pairwise registrations to exhaustively evaluate the outlier filter solutions.
These registrations  were computed offline using Compute Canada's Supercomputers. 
In the second set of experiments (\autoref{sssec:husky}), we collected data with a mobile robot on an indoor-outdoor trajectory. 
We then tested on this trajectory the two best performing filters (\autoref{sec:registration}) for a real-time mapping task (\autoref{fig:husky}).

\vspace{-5pt}
\subsection{Pairwise Outlier Filter Tests}
\label{sec:registration}
\begin{figure}[htbp]
\centering
\vspace*{-10pt}
\includegraphics[width=0.47\textwidth]{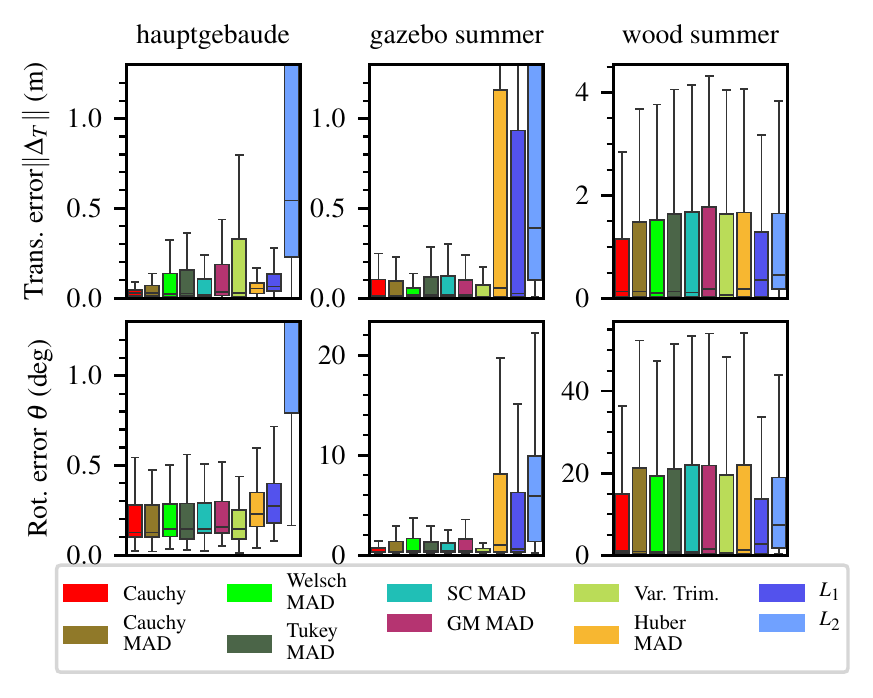} %
\caption{
	Histogram of the performance of each outlier filter for the parameter value with lowest median error in that particular environment.
    }
\label{fig:histogram}
\vspace*{-17pt}
\end{figure}

Pairwise registration results are broken down into three parts. We first discuss the performance (median error) of each filter as a function of its parameter, for all environments compounded (\autoref{fig:param_infl}).
We then closely analyze the distribution of errors for the best parameter values found from this search, for the three environments (\autoref{fig:histogram}).
In the third part, we evaluate how robust the best parameters are to a change in the environment (\autoref{tab:results}).

\subsubsection{Parameter Search over all Environments Compounded}
\label{sssec:A}
We computed the error metrics $\norm{\bm{\Delta}_T}$ for all registrations, irrespective of the environment. \autoref{fig:param_infl} shows the median translation error for M-estimators, as a function of the filter's tuning parameter value. 
We can see that all filters with parameters have a single global minimum, located in a relatively flat valley. Most solutions have similar best performance, while \emph{Huber MAD} slightly under-performing. 


\textbf{Comparison with $L_2$ (Vanilla ICP):} For large values of configuration parameter $k$, all filters (excepted \emph{GM}) performed similarly to $L_2$ (top dashed line), i.e. outlier filtering is effectively disabled.
This is unsurprising, as these M-estimators have the property that $\lim_{k \to \infty} w(e, k) \approx 1$. For small values of $k$, the performance depends on the estimator used. It can nevertheless be categorized into two trends.
For three filters (\emph{Huber MAD}, \emph{Cauchy} and \emph{Cauchy MAD}), the translation error degrades smoothly towards $L_2$ as $k\rightarrow 0$, without surpassing it.
For the four others, performance can become \emph{much worse} than $L_2$ with $k\rightarrow 0$, although for \emph{Welsch} and \emph{Tukey} the performance eventually goes back down to $L_2$.
From all this, we can conclude that: \emph{1)} it is preferable to overestimate the parameter for all M-estimator than to underestimate it, to avoid too much rejection of inliers; \emph{2)} peak performance varied little from one filter to another (less than \SI{2}{\cm}), except for \emph{Huber MAD}; and \emph{3)} \emph{Huber MAD}, \emph{Cauchy} and \emph{Cauchy MAD} never performed worse than $L_2$.


\textbf{Comparison with $L_1$:} Despite being unsophisticated, $L_1$ almost always outperform all outlier filters approach, except for a narrow band of parameter values. \emph{Huber} is the only M-estimator that does not outperform $L_1$. 
However, as we will see later in \autoref{fig:histogram}, $L_1$ exhibits a much greater variance in certain environments, which could be problematic if one is interested in minimizing the risk of being lost.

\textbf{Other observations:} \emph{Cauchy} and \emph{Cauchy MAD} have the same curve, but with an offset, with \emph{Cauchy MAD}'s minimum centered at $k=1$ by the auto-scaler. 
Since \emph{GM} and \emph{SC} shares the same function, they have similar curve, differing only by a slight offset.
\emph{Welsch} and \emph{Tukey} also have similar curve with an offset, despite having very different weight functions $w(e)$. 

\subsubsection{Distribution of errors for the best fixed filter parameter, for three environments}
\label{sssec:B}
In \autoref{fig:param_infl} we showed only the median error, for all environments combined together. 
As such, this median does not tell the whole story.
We thus computed in \autoref{fig:histogram} the distribution of errors of all filtering approaches, for three different environments. 
Importantly, the filter parameters were fixed in these experiments, and were established from the best results found from \autoref{fig:param_infl}.
From these experiments, the environment influence on the registration is clear: the structured one (\emph{hauptgebaude}) is noticeably easier in rotation, while the unstructured one (\emph{Wood Summer}) is by far the hardest in rotation and in translation. 
Furthermore, all error distributions are asymmetric (heavy-tailed). 
Apart from the unstructured environment of \emph{wood summer}, all outlier filters have error spreads and median errors significantly smaller than $L_2$. This further confirms the lack of robustness of $L_2$. 
Although $L_1$ has a favorable median error in the semi-structured environment \emph{gazebo summer}, its error spread is much greater that most other M-estimators.
\emph{Huber} performed even worst than $L_1$ in that environment despite being a "robust" version of the former. Finally, we observed that \emph{Cauchy} has an error spread noticeably smaller than its auto-scaled counterpart (\emph{Cauchy MAD}).

\subsubsection{Robustness of best fixed parameter across Environments}
\label{sssec:C}
To determine if a filter is robust across environment changes, we used the following metric: if the best parameter of a filter for each three environments are all within the global flat valley, this filter is considered robust.
We define this global flat valley as the range of parameter that perform better than $L_1$, when comparing the median translation error.
For instance, \autoref{fig:param_infl} show the best parameters of \emph{Cauchy} for three environments.
Since all three parameter values have an error below $L_1$, \emph{Cauchy} is robust as per our metric. 

In \autoref{tab:results}, the best performing parameters for all filters are presented, for 3 environments. The best overall outlier filter for these experiments is \emph{Var. Trim}.
It has half of the error of the second best filter in \emph{Wood Summer}.
It is also the only hard rejection filter that met our robustness criteria describe above. 
Filters such as \emph{Cauchy}, \emph{Cauchy MAD}, \emph{Welsch MAD}, and \emph{Tukey MAD} are also robust across our environment changes, as they all have values below $L_1$ (indicated by bold notation).
On the subject of auto-scaling, \emph{Cauchy Berg} is out performed by \emph{Cauchy} and \emph{Cauchy MAD} for all environment except \emph{Wood Summer}.
The optimal parameter of \emph{Welsch} for two environments  is $k=2$, the exact value for \ac{RQE} kernel function (as discussed in \autoref{ssec:crisp}), which means that our experiments agree with the theory around kernel function. 
The parameters of \emph{Max distance} and \emph{Trim} depend on the environment, possibly indicating a lack of robustness. We can attribute this to the fact that they are hard rejection, and that they lack the adaptability of \emph{Var. Trimmed}.

\begin{table}
\setlength\tabcolsep{4.5pt}
	\centering
	\caption{
		All 14 outlier filters were tested on the 3 environments and for a variety of parameter values.
        The parameter corresponding to the smallest median error for each dataset is shown.
        In the median error columns, the outlier filter with the best performance is in bold.
        In the parameter columns, if the parameter value is located inside the global flat valley, then it is in bold. 
        If all three environments are within the valley, then 'All' is in bold.
	}
	\begin{tabu} {@{}X|cccc|cccc@{}}
		\toprule
		Outlier Filters & \multicolumn{4}{c|}{Median Error (mm)} & \multicolumn{4}{c}{Parameters 
        
     } \\
		&     H & GS & WS &    All & H & GS & WS &    All \\
\midrule
$L_2$              &               544 &           390 &          459 &          409 &            n/a &            n/a &            n/a &   n/a \\
$L_1$              &                66 &            25 &          350 &           68 &            n/a &            n/a &            n/a &   n/a \\
Huber MAD          &                54 &            58 &          176 &           84 &           0.05 &           0.33 &           0.67 &  0.33 \\
Cauchy             &                28 &            11 &          131 &           37 &  \textbf{0.05} &  \textbf{0.10} &  \textbf{0.32} &  \textbf{0.20} \\
Cauchy MAD         &                28 &            14 &          132 &  \textbf{28} &  \textbf{0.40} &  \textbf{1.00} &  \textbf{1.59} &  \textbf{0.80} \\
Cauchy Berg        &                39 &            18 &           99 &           39 &  \textbf{0.01} &  \textbf{0.00} &           0.05 &  0.01 \\
SC MAD &       \textbf{21} &            19 &          111 &           31 &           0.50 &  \textbf{2.53} &  \textbf{3.18} &  1.00 \\
GM MAD  &                34 &            19 &          180 &           47 &           1.08 &  \textbf{4.52} &          11.72 &  4.52 \\
Welsch MAD         &                23 &            17 &          109 &           36 &  \textbf{2.00} &  \textbf{2.00} &  \textbf{3.18} &  \textbf{1.59} \\
Tukey MAD          &                26 &            19 &          130 &           34 &  \textbf{2.53} &  \textbf{5.04} &  \textbf{6.35} &  \textbf{3.18} \\
Student            &                40 &            32 &          178 &           60 &           0.10 &           0.13 &           1.37 &  0.16 \\
Max. Dist.           &                45 &            38 &          112 &           58 &           0.30 &  \textbf{0.40} &           0.60 &  0.40 \\
Trim               &                39 &            24 &          284 &           63 &           0.63 &  \textbf{0.68} &           0.89 &  0.68 \\
Var. Trim.         &                28 &    \textbf{9} &  \textbf{58} &  \textbf{27} &  \textbf{1.91} &  \textbf{2.35} &  \textbf{2.35} &  \textbf{1.91} \\
\bottomrule
	\end{tabu}
    \vspace{2pt}
    \footnotesize{Legend: H = Hauptgebaude, GS = Gazebo Summer, WS = Wood Summer} 
    \label{tab:results}
    \vspace{-20pt}
\end{table}

\vspace{-5pt}
\subsection{Test on a full Indoor-Outdoor Trajectory}
\label{sssec:husky}
In \autoref{fig:husky}, two filters (\emph{Cauchy MAD} and \emph{Var. Trim.}) that were tested in a challenging route containing both indoor and outdoor portions. 
These filters were chosen for their good performance in our offline tests.
The robot used a Velodyne HDL-32e for scanning and used a mix of wheel encoder and an Xsens MTi-30 IMU for odometry.
To estimate the trajectory, an \ac{ICP}-based SLAM was used with a moving window map.
The moving window map was create by randomly decimating the map's point.
For ground truth, the \ac{ICP}-based SLAM was done offline without moving windows.
The outlier filter performance was evaluated by comparing the cumulative registration error to the ground truth.
Multiple parameter values were tested, all near the best value from~\autoref{tab:results}.
For \emph{Var. Trimmed}, all parameter values had difficulties in the highlighted zone A in \autoref{fig:husky}, which happen in a small corridor.
This confirmed that \emph{Var. Trimmed} has problem in structured environments as in \autoref{fig:histogram}-\emph{Hauptgebaude}. 
For \emph{Cauchy MAD}, all tested parameters finished close to the ground truth. The parameter with the best performance was $k = 3.0$, which demonstrate that the valley for this experiment is shifted to higher $k$ values than our offline tests.
We think that this change is caused by the different sensors used, and because this test is a registration between scan to (small) map, while our offline tests were between two scans. 


\vspace*{-5pt}
\section{Conclusion}
\vspace*{-1pt}
\label{conclusion}
In this paper, we performed exhaustive robustness experiments on a wide range of outlier filters, in the context of \ac{ICP}. After analysis, we concluded that all robust solutions have similar performances, with a number of particularities worth noting. For instance, $L_1$ exhibit good overall performance, despite having no parameter. Also, using MAD as auto-scale improves accuracy, but does not relieve from parameter tuning. When appropriately tuned, \emph{Var. Trim.} has the best accuracy, with a translation error under \SI{27}{\mm}, while the best M-estimator is \emph{Cauchy MAD} with roughly the same error. 
Moreover, we demonstrated the necessity of a well-tuned outlier filter for robust registration. In particular, \emph{Welsch}, \emph{Tukey}, \emph{GM}, and \emph{SC} should be employed carefully, as they have the potential to produce worse estimate than $L_2$ when mis-tuned. This fact should be kept in mind when assessing the risks associated with parameter selection.

Encouraged by the result of $L_1$, further investigation will be made to adapt registration solution related to $L_p$ norms \citep{Bouaziz2013a} into the standard \ac{ICP} pipeline.
Furthermore, the link between \ac{RQE}, a Gaussian kernel and \emph{Welsch} opens the door to a family of kernels to be studied for outlier filters along with an integration of the kernel correlation solution \citep{Kanade2004} and EM-ICP \cite{Granger2002} into a generic version of \ac{ICP}.





\printbibliography

\end{document}